# VECTORIZACIÓN SEMI-AUTOMÁTICA DE REDES LINEALES EN MAPAS CARTOGRÁFICOS EN COLOR

C. Miravet, E. Coiras, J. Santamaría.
División Aeroespacial
SENER Ingeniería y Sistemas, S.A.
Severo Ochoa 4, 28760, Tres Cantos (Madrid).

**RESUMEN**

Se presenta un sistema semi-automático de vectorización de redes de objetos lineales (carreteras, ríos, etc.) en mapas cartográficos digitalizados. En este sistema, la intervención humana queda reducida a la selección gráfica interactiva de los atributos de color de la información a obtener. Con estos datos, el sistema realiza una extracción preliminar de la red lineal, que se completa, refina y vectoriza mediante un procedimiento automático. Se presentan resultados de la aplicación del sistema sobre imágenes digitalizadas de mapas de distinta procedencia y escala.

PALABRAS CLAVE: Vectorización de mapas, segmentación en color interactiva, procedimientos de enlace de contornos.

**ABSTRACT**

A system for semi-automatic vectorization of linear networks (roads, rivers, etc.) on rasterized cartographic maps is presented. In this system, human intervention is limited to a graphic, interactive selection of the color attributes of the information to be obtained. Using this data, the system performs a preliminary extraction of the linear network, which is subsequently completed, refined and vectorized by means of an automatic procedure. Results on maps of different sources and scales are included.

## INTRODUCCIÓN

En el ámbito del tratamiento informático de la información cartográfica, se entiende por vectorización el análisis y extracción estructurada de la información contenida en un mapa previamente digitalizado (generalmente mediante el uso de un *scanner*) y su almacenamiento en un formato vectorial, basado en primitivas, adecuado a su utilización en aplicaciones del tipo de las proporcionadas por los sistemas de información geográficos (GIS). Las implicaciones económicas de la obtención de un sistema semi-automático de estas características son importantes debido a los siguientes factores:

- Uso creciente de los sistemas de tipo GIS.

- Existencia de una base de datos ingente de mapas en soporte papel cuyo contenido no se encuentra disponible en formato vectorial.

- Laboriosidad asociada al procedimiento de vectorización manual de mapas (mediante tableta digitalizadora).

En el presente trabajo se describe una aplicación *software* de vectorización semi-automática de objetos constituidos por redes lineales (carreteras, redes fluviales, etc.) a partir de mapas cartográficos en color, digitalizados, presentándose resultados de su aplicación sobre mapas de distinta procedencia y escala. El uso de este entorno permite reducir el tiempo y laboriosidad asociados al proceso de vectorización, mediante la obtención, con mínima intervención humana, de una primera versión del resultado. En la fase final del proceso, esta primera versión puede retocarse de forma manual, mediante operación sobre un interfaz gráfico adecuado. Este retoque se limita típicamente a zonas específicas de la imagen, dada la alta calidad global de los datos suministrados por el sistema automático.

A continuación se presenta una descripción básica por módulos del procedimiento desarrollado, finalizándose con la presentación de varios ejemplos de aplicación del sistema.

## DESCRIPCIÓN DEL PROCEDIMIENTO

El sistema desarrollado proporciona un entorno gráfico que incorpora un procedimiento básico de vectorización semi-automática de redes de elementos lineales (carreteras, redes fluviales, etc.) a partir de mapas cartográficos en color, digitalizados. En estos mapas, los colores que codifican elementos diferentes de información presentan una separación, en términos perceptuales, suficiente para posibilitar la interpretación visual de la información presentada. En el método desarrollado, este hecho se utiliza de forma directa, al realizarse la selección de la información a extraer mediante especificación directa de su rango de colores asociado. Para ello, se genera una representación 2D de la frecuencia de aparición de colores en la imagen, o histograma 2D. La utilización de este histograma permite determinar de forma visual e interactiva el rango de colores correspondiente a la información de interés, mediante selección de un intervalo de valores sobre el histograma de color 2D y observación de la información extraída correspondiente a esos atributos de color. El proceso de obtención del histograma 2D se describe en detalle en el apartado siguiente.

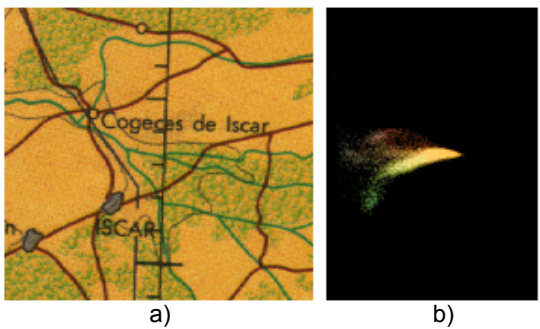

**Figura 1**: a) mapa en color; b) histograma 2D

En la figura 1 se presenta un fragmento de un mapa en color junto con el histograma 2D correspondiente. En este histograma, los diversos colores aparecen con su tono y saturación original y con una intensidad proporcional a su frecuencia de ocurrencia en la imagen analizada, obteniéndose distintos cúmulos correspondientes a cada uno de los colores presentes en la escena. Esto permite al operador del sistema la extracción preliminar de la información requerida mediante la simple delimitación gráfica sobre el histograma del cúmulo correspondiente a los objetos de interés.

En la figura 2 se presenta el histograma 2D sobre el que se ha realizado la selección manual de un cúmulo de color, junto con el resultado de la extracción de las zonas del mapa correspondientes. Como puede observarse, si bien el procedimiento de extracción preliminar permite obtener la mayor parte de la información disponible correspondiente al rango de colores seleccionado, esta viene afectada por diversos artefactos (aparición de zonas ruidosas, pérdidas de conectividad en la malla, obtención de elementos de anchura variable, etc.) que impiden su uso directo en aplicaciones que requieran un producto final de alta calidad. Por ello, la fase final del procedimiento está dedicada a completar y depurar los resultados de la extracción preliminar, habiendo sido diseñado un método automático compuesto por los siguientes pasos básicos:

- Aplicación de un procedimiento de crecimiento de regiones con semilla dada por los resultados de la extracción preliminar.
- Eliminación de regiones pequeñas asociadas a fenómenos de ruido.
- Representación de las regiones lineales extraídas por su línea media, mediante aplicación de procedimientos morfológicos (esqueleto).
- Descripción, en forma de grafo, de las regiones obtenidas.
- Eliminación de cadenas/ramas con longitud menor a un umbral.
- Aproximación poligonal de la red lineal extraída.
- Restauración de pérdidas de conectividad en la red.
- Almacenamiento de la red lineal en un formato vectorial adecuado.

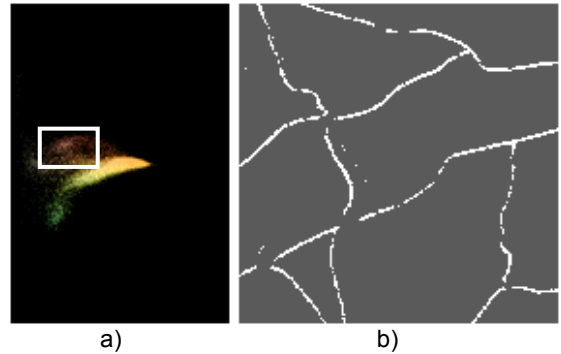

**Figura 2**: a) selección sobre histograma; b) extracción preliminar de información

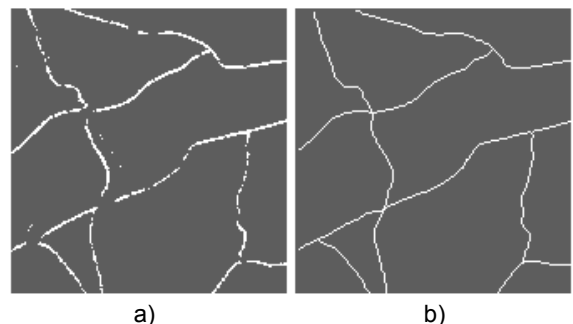

**Figura 3**: a) extracción preliminar; b) resultado final del procedimiento

En la figuras 3 y 4 se presentan, respectivamente, el resultado global y por fases de este proceso sobre la imagen test. En los apartados siguientes se presentan en mayor detalle los pasos del procedimiento implementado.

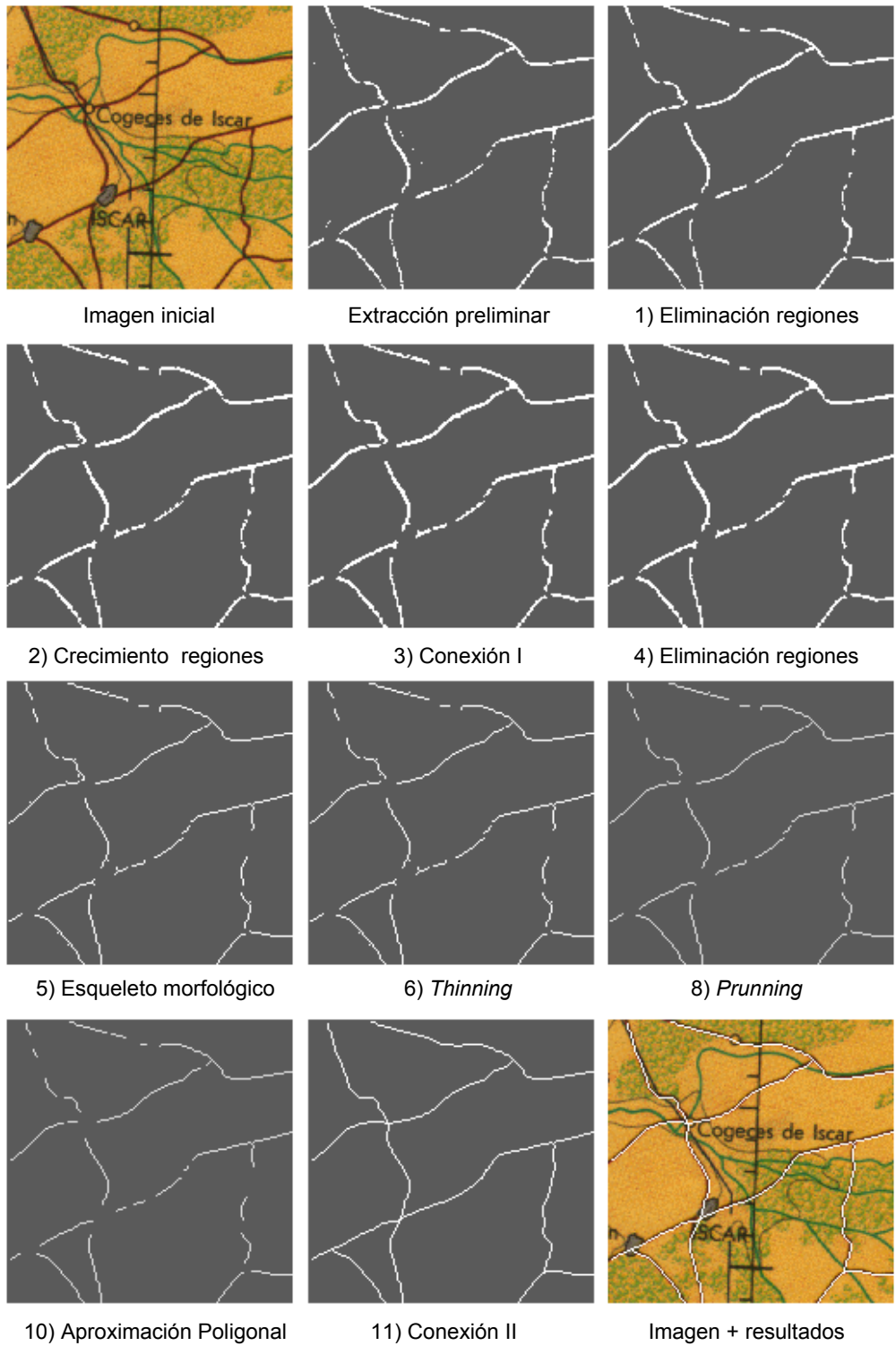

**Figura 4**: Fases principales del procedimiento de refinamiento de la información (ver texto).

## PASOS DEL PROCEDIMIENTO

### Cálculo del histograma de color. Extracción preliminar de la información

El objetivo de la primera fase del proceso es la obtención y representación en forma de imagen (histograma 2D) de la información relativa a las frecuencias de ocurrencia de los distintos colores presentes en el mapa a examinar. El número reducido de colores utilizado frecuentemente en los mapas cartográficos, junto con la uniformidad de sus atributos de color, llevará a la aparición de acumulaciones claramente diferenciadas en esta representación de frecuencias. La interfaz gráfica permitirá encuadrar la zona correspondiente a cada cúmulo, operación que proporcionará de forma directa el rango de atributos de color de la información asociada del mapa, permitiendo su extracción preliminar.

La determinación del histograma 2D puede descomponerse en los siguientes pasos de proceso:

a) *Especificación del espacio de representación del color*

La selección del espacio de representación a utilizar tendrá una influencia determinante en la distribución de las acumulaciones correspondientes a los distintos colores utilizados en el mapa, debiendo seleccionarse aquel espacio que provea una mejor discriminación entre las distribuciones típicas de color en el conjunto de casos patrón utilizados.

En particular, en nuestro caso se consideraron los siguientes espacios de representación [Pratt, 1991], [Jain, 1989]:

- Espacio HSI (tono-saturación-intensidad).
- Espacio $L^*u^*v^*$.

Para la aplicación considerada, el espacio HSI es el que ha proporcionado mejores resultados, siendo el incorporado al procedimiento. Este espacio esta definido por las siguientes ecuaciones de transformación [Pratt, 1991]:

$$\begin{pmatrix} I \\ V_1 \\ V_2 \end{pmatrix} = \begin{pmatrix} \frac{1}{3} & \frac{1}{3} & \frac{1}{3} \\ \frac{-1}{\sqrt{6}} & \frac{-1}{\sqrt{6}} & \frac{2}{\sqrt{6}} \\ \frac{1}{\sqrt{6}} & \frac{-2}{\sqrt{6}} & 0 \end{pmatrix} \cdot \begin{pmatrix} R \\ G \\ B \end{pmatrix}$$

$$H = tan^{-1}\left(\frac{V_2}{V_1}\right) \qquad [1]$$

$$S = \left(V_1^2 + V_2^2\right)^{1/2}$$

Donde (*R, G, B*) y (*H, S, I*) son, respectivamente, las coordenadas de los espacios de representacion inicial y final. $V_1$ y $V_2$ son variables auxiliares utilizadas en el proceso.

b) *Proyección del espacio de color*

Una vez seleccionado el espacio de representación 3D del color a utilizar, éste debe proyectarse a un subespacio bidimensional a fin de permitir una correspondencia directa con las coordenadas geométricas del histograma 2D. De forma general, los distintos tipos de información se codifican principalmente mediante variaciones de tono y saturación, incorporándose poca información adicional en la intensidad. Por ello, se utilizará de forma prioritaria la proyección sobre el espacio S(saturación, eje X) – H (tono, eje Y).

Una excepción importante la constituye la extracción de objetos representados en una tonalidad de gris. En este caso, el valor de tono resulta irrelevante, siendo mas apropiada la proyección sobre el espacio S(saturación, eje X) – I (intensidad, eje Y). En esta representación, las tonalidades de gris pueden separarse del resto de colores en función de su menor valor de saturación. La tonalidad a extraer puede, entonces, seleccionarse mediante especificación del rango de intensidades asociado.

c) *Generación del mapa de frecuencias de ocurrencia*

Una vez seleccionada la operación de proyección adecuada a las características de la información a extraer, se aplican las operaciones anteriores sobre cada punto de la imagen de entrada, realizándose el computo de las frecuencias de ocurrencia de los distintos valores proyectados (X, Y).

d) *Generación del histograma 2D*

En la última fase del proceso se incorpora de forma simultanea al histograma 2D la información de las frecuencias de ocurrencias de las distintas gamas de colores junto con las características crómaticas suficientes que permitan la identificación visual de estos colores.

En el caso de la extracción de objetos en color, que ha sido el considerado de forma prioritaria, esto se ha realizado de forma directa mediante la generación de una imagen en color de dimensiones idénticas al mapa de frecuencias de ocurrencia y con las siguientes coordendas de color HSI:

- Intensidad dada por el logaritmo de la frecuencia de ocurrencia, debidamente normalizado.

- Saturación dada por el valor de la coordenada X (saturación discretizada en la imagen original) si este valor es compatible con el valor de intensidad.

- Tono dado por el valor de la coordenada Y (tono discretizado en la imagen original).

Como resultado de la operación anterior, se obtiene un mapa de distribución de los colores presentes en la imagen analizada, con intensidades dadas por la frecuencia de ocurrencia. Los colores no presentes en la imagen original aparecen en negro.

La disposición de este mapa de distribución permite la selección visual de los rangos de color asociados a cualquier color presente en la imagen inicial, permitiendo la extracción preliminar de la información asociada. Esta información se completa y depura de forma automática en la segunda fase del proceso.

**Refinamiento de la información extraída**

Como se ha mencionado anteriormente, la información obtenida como resultado de la extracción preliminar presenta frecuentemente diversos artefactos que impiden el uso directo en aplicaciones que requieran un producto final de alta calidad. Por ello, esta fase del procedimiento se dedica a la eliminación de estos artefactos mediante aplicación de un proceso automático de refinamiento. Como resultado de este proceso se obtiene una red de elementos lineales, descrita de forma interna como un grafo, que permite su almacenamiento directo en el formato vectorial de elección.

A continuación se enumeran los pasos de proceso del algoritmo, junto con la referencia del apartado donde se detalla cada paso.

1) Eliminación de zonas extraídas con área menor a un umbral, $T_1$. (apartado *a*)

2) Aplicación de un algoritmo de crecimiento de regiones sobre las zonas extraídas. (*b*)

3) Conexión de las zonas adyacentes. (*c*)

4) Eliminación de zonas con área menor a un umbral, $T_2$. Eliminación de regiones internas a las zonas extraídas, con área menor a un tercer umbral, $T_3$.

5) Determinación del esqueleto morfológico de las zonas extraídas. (*d*)

6) Aplicación de un proceso de *thinning* morfológico. (*e*)

7) Conversión de la imagen extraída a grafo. (*f*)

8) Eliminación, *prunning*, de cadenas y ramas con longitudes inferiores, respectivamente, a L1 y L2. (*g*)

9) Eliminación de puntos extremos de las cadenas. (*h*)

10) Aproximación poligonal de la red lineal obtenida. (*i*)

11) Restauración de pérdidas de conectividad en la red. (*j*)

12) *Prunning* del grafo con umbrales menos restrictivos a los utilizados en *8)*.

13) Almacenamiento vectorial. (*k*)

En la figura 4 se presentan los resultados de la aplicación de las fases principales de este algoritmo sobre las regiones extraídas de la figura 2. En los apartados siguientes se presentan en mayor detalle los distintos pasos de proceso. Se concluye con la presentación de los resultados obtenidos mediante aplicación de método.

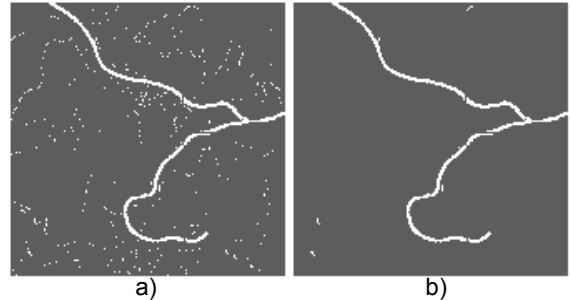

**Figura 5**: a) zonas extraídas; b) eliminación de regiones basada en área

*a) Eliminación de regiones con área menor a un umbral*

El algoritmo implementado determina el conjunto de regiones presentes en la imagen de entrada, caracterizadas como conjuntos conexos de pixels con nivel de gris uniforme. Estas regiones se obtienen mediante el uso del algoritmo de búsqueda de componentes conexos para grafos de Rosenfeld y Platz [Haralick y Shapiro, 1992], determinándose de forma simultanea las áreas de cada una de las regiones encontradas y las relaciones de adyacencia entre regiones. Con esta información, el algoritmo elimina regiones con área inferior a un umbral, $T_x$, o regiones internas con área inferior a un segundo umbral, $T_y$. En la figura 5 se muestra el resultado de aplicar el algoritmo de eliminación de regiones sobre los resultados de una extracción preliminar.

*b) Crecimiento de regiones*

La aplicación de un algoritmo de crecimiento de regiones permite completar el perfil de las regiones extraídas mediante incorporación de pixels vecinos con atributos de color similares a los de estas zonas. El método implementado es una extensión para imágenes en color del algoritmo de crecimiento jerárquico de regiones descrito en [Adams y Bischof, 1994]. En esta extensión, se utiliza la siguiente métrica de diferencias de color, d, [Fuertes et al., 1996], entre dos puntos del espacio HSI con coordenadas de color ($H_1$, $S_1$, $I_1$) y ($H_2$, $S_2$, $I_2$), respectivamente:

$$d = \left(d_I^2 + d_C^2\right)^{1/2}$$
$$d_I = |I_1 - I_2|$$
$$d_c = \left(S_1^2 + S_2^2 - 2S_1S_2 \cos\phi\right) \qquad [2]$$
$$\phi = \begin{cases} |H_1 - H_2|, & si\ |H_1 - H_2| < \pi \\ 2\pi - |H_1 - H_2|, & si\ |H_1 - H_2| > \pi \end{cases}$$

Una vez aplicado el algoritmo de crecimiento de regiones, el método calcula y almacena los valores medios de las coordenadas RGB sobre las zonas extraídas. Estos valores definirán el color asignado a las zonas proporcionadas como resultado final del proceso.

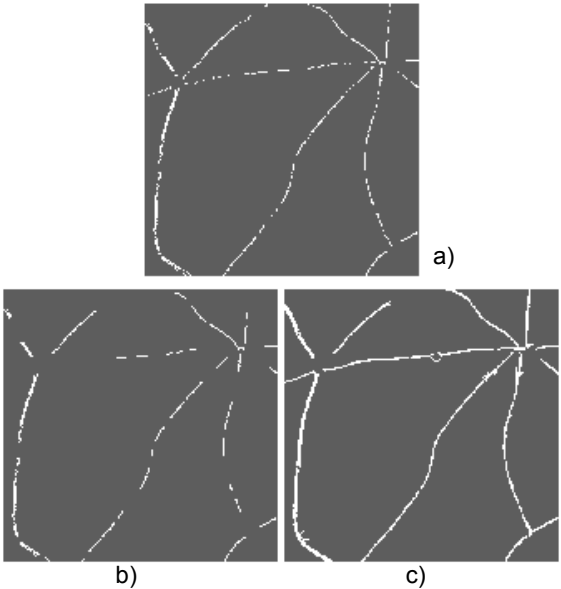

**Figura 6**: a) zonas extraídas; b) eliminación de regiones basada en área; c) Resultado del crecimiento de regiones.

En la figura 6 se muestra el resultado de la aplicación del método sobre una imagen de entrada. Como puede observarse, la aplicación del método permite recuperar la información de regiones fragmentadas del objeto que han sido descartadas, junto con las zonas ruidosas, en la eliminación de regiones basada en área.

*c) Conexión preliminar de regiones adyacentes*

En este paso de proceso se procede a una primera conexión de regiones que han sido desunidas indebidamente durante el proceso de extracción preliminar. Esta conexión se realiza mediante la activación de puntos del fondo que estén conectados (según el criterio de vecindad-4) a dos o mas puntos previamente activados. El método de unión utilizado sigue deliberadamente un criterio conservador, dejando la carga principal de la conexión de ramas separadas al algoritmo detallado en la sección *j*).

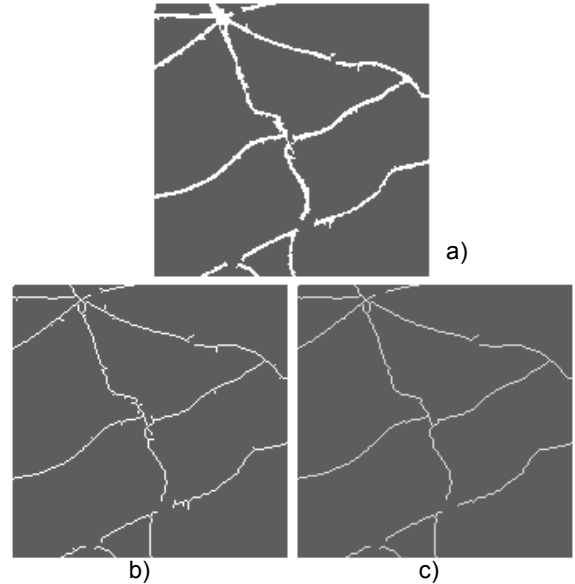

**Figura 7**: a) Resultado crecimiento de regiones; b) esqueleto morfológico; c) *Prunning*.

*d) Determinación del esqueleto morfológico*

En este paso del proceso, se extrae el esqueleto morfológico de las regiones obtenidas como resultado de los pasos anteriores. Este algoritmo reduce las regiones obtenidas a su línea media, con espesor de un pixel.

En la literatura existen diversos procedimientos de extracción del esqueleto. En nuestro sistema se ha utilizado un método robusto basado en la distancia de *chamfer* [Borgefors, 1988], de acuerdo a lo descrito en [Sanniti di Baja, 1994].

En la figura 7 b) se muestra el esqueleto de las regiones extraídas tras aplicación del algoritmo de crecimiento de regiones.

*e) Algoritmo de adelgazamiento (thinning) morfológico*

Sobre los resultados de la extracción del esqueleto morfológico se aplica una operación adicional de adelgazamiento basado en el algoritmo de Rutovitz [Jain, 1989]. Este paso asegura la obtención de una cadena de espesor unidad en toda su extensión, reduciendo la complejidad de su representación en forma de grafo en el paso inmediatamente posterior del procedimiento. Las modificaciones ocasionadas por este algoritmo se reducen típicamente a la corrección unos pocos pixels por imagen, donde la obtención del esqueleto no había conducido a una cadena de espesor unidad.

f) *Conversión de la red lineal a grafo*

Una vez reducida la red lineal a un conjunto de arcos de espesor unidad, esta malla se representa en forma de grafo [Chartrand y Oellermann, 1993]. En la representación elegida, los nodos del grafo están constituidos por puntos singulares de la malla: puntos extremos, bifurcaciones y puntos aislados. Los enlaces entre nodos representan los distintos arcos constituyentes de la malla.

En nuestro procedimiento, se han utilizado estructuras de datos basadas en el concepto de lista enlazada, lo que ha permitido una flexibilidad máxima en operaciones de manipulación tales como la creación, eliminación y fusión de distintos elementos del grafo (nodos y arcos).

g) *Podado (prunning) del grafo*

Una vez obtenido el grafo, se aplica un operación de podado (*prunning*) mediante la cual se eliminan las cadenas y ramas con longitudes inferiores a dos umbrales seleccionados por el usuario. La operación implementada mantiene la integridad del grafo, al tener en cuenta las modificaciones, desaparición de bifurcaciones, asociadas al proceso de eliminación de ramas.

En la figura 7 c) se muestra el resultado de la aplicación del método sobre el esqueleto morfológico de una red lineal.

h) *Eliminación de puntos extremos de las cadenas*

En este paso del proceso se eliminan los *n* puntos mas próximos a los extremos en los arcos abiertos. Estos puntos acumulan frecuentemente los mayores errores de localización, especialmente en los casos de pérdida de conectividad por intersección de la red con elementos del mapa con atributos de color similares, pudiendo dar lugar a la aparición de segmentos espurios durante el proceso de aproximación poligonal.

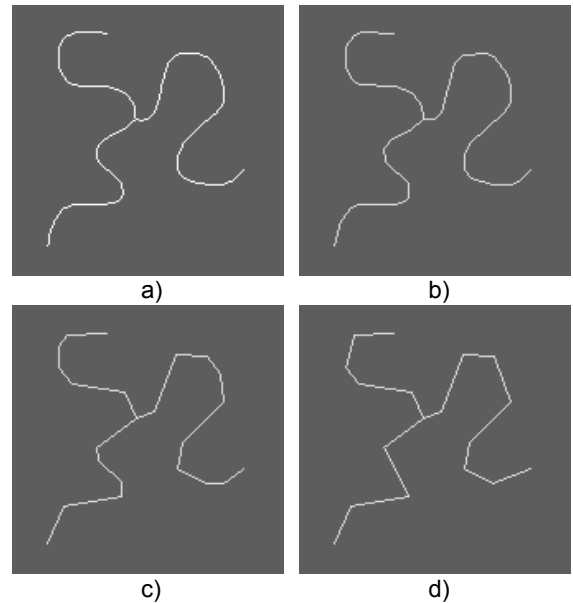

**Figura 8**: Aproximación poligonal de la red. a) red original; Aproximaciones poligonales con precisiones: b) 1 pixel; c) 3 pixels; d) 5 pixels

i) *Aproximación poligonal*

En este paso del proceso se lleva a cabo la aproximación de los arcos obtenidos por un conjunto de segmentos lineales, mediante aplicación del algoritmo de Forsen de ajuste iterativo de puntos extremos [Duda y Hart, 1973].

Este algoritmo permite ajustar el nivel de precisión deseado en el proceso de aproximación lineal, mediante especificación de la distancia máxima admisible entre un punto cualquiera del arco y la aproximación obtenida. En la figura 8 se presenta un fragmento de red junto con las aproximaciones poligonales correspondientes a errores máximos de 1, 3 y 5 pixels. Las aproximaciones lineales correspondientes contienen, respectivamente, 40, 22 y 17 segmentos.

j) *Restauración de pérdidas de conectividad en la red*

La aplicación de este paso permite reparar de forma automática las pérdidas indebidas de conectividad en la malla, ocasionadas por fenómenos tanto de ruido como de interferencia con elementos de información próximos (cuadricula, texto, otras redes, etc.).

El procedimiento desarrollado está basado en la aplicación secuencial de los siguientes pasos de proceso:

*j.1) Extracción de los puntos extremos del grafo.*

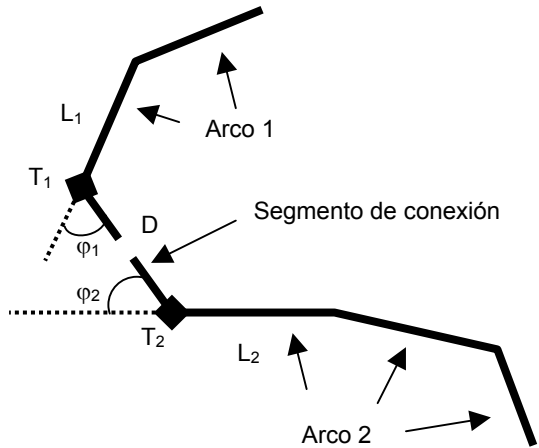

**Figura 9**: Parámetros geométricos en la conexión de dos arcos. $T_{1,2}$: extremos; D: distancia de conexión; $\varphi_{1,2}$: ángulos del segmento de conexión con el segmento final del arco; $L_{1,2}$: longitudes de los segmentos finales de los arcos.

*j.2) Determinación inicial de la viabilidad de conexión entre cada par de puntos extremos extraídos.*

En el sistema desarrollado, esta determinación se realiza exclusivamente en función de parámetros geométricos de la conexión, en particular (ver figura 9):

- Distancia euclidea entre puntos extremos, *D*.

- Ángulos, $\varphi_{1,2}$, formados por el segmento de conexión entre extremos con los segmentos finales de la aproximación poligonal de cada arco.

- Longitud, $L_{1,2}$, de los segmentos finales de cada arco. Estas magnitudes se utilizan como medidas cuantitativas del nivel de confianza asignable al valor de la dirección en el extremo del arco.

En una segunda fase se prevé la incorporación de criterios de constancia de color a lo largo de la conexión. En cualquier caso, debe tenerse en cuenta que estos criterios no podrán ser usados de forma directa en las zonas donde la pérdida de conectividad se deba a la intersección de la red con otros elementos del mapa.

En función de los parámetros geométricos utilizados, dos puntos extremos se considerarán conectables si:

1. El coste de conexión es menor que un umbral, $T_D$. Este coste viene dado por la distancia euclidea (D) entre ambos extremos modulada por una función de los parámetros angulares ($\varphi_{1,2}$) del problema.

2. La longitud de los últimos segmentos lineales de ambos arcos, $L_{1,2}$, es mayor que un umbral, $T_L$.

3. El mínimo ángulo formado por el segmento de conexión con los segmentos finales de los arcos a unir es menor que un umbral, $T_{ANG\_MIN}$. Es decir, el segmento de conexión representa una extensión natural de al menos uno de los arcos a unir.

4. El máximo ángulo formado por el segmento de conexión con los segmentos finales de los arcos a unir es menor que un segundo umbral, $T_{ANG\_MAX}$. Es decir, el segmento de conexión no representa una extensión improbable de ninguno de los arcos a unir.

5. Las relaciones entre la distancia de conexión y las magnitudes $L_{1,2}$ se encuentran dentro de unos rangos preestablecidos.

*j.3) Inhibición selectiva de conexiones entre pares de puntos extremos.*

En este paso se realiza una selección de las conexiones determinadas en el paso anterior en función de los siguientes criterios:

1. Se eliminan aquellas conexiones entre arcos que intersecten otros arcos de la red.

2. Se eliminan aquellas conexiones que den lugar a ciclos cerrados excesivamente elongados o de longitud inferior a un umbral. La verificación de este criterio requiere la disponibilidad de un método de cálculo de distancias mínimas entre puntos extremos de la red, para lo que se ha utilizado el algoritmo de Dijkstra de búsqueda de caminos mínimos entre nodos de grafos con peso [Chartrand y Oellermann, 1993].

En la figura 10 se presentan dos ejemplos de conexiones eliminadas mediante aplicación de los criterios inhibitorios mencionados.

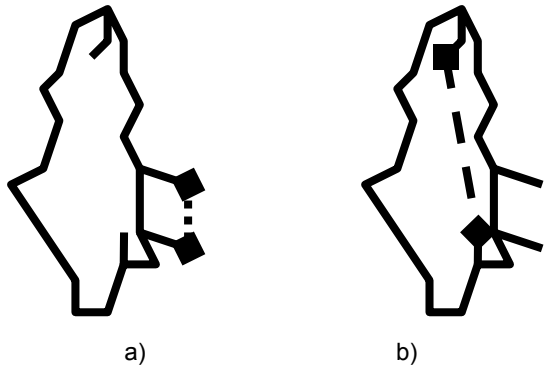

a) b)
**Figura 10**: Criterios de inhibición de conexiones. a) conexión (línea de puntos) entre extremos (rombos) descartada al dar lugar a un ciclo cerrado con longitud por debajo de un umbral; b) conexión descartada al dar lugar a un ciclo cerrado excesivamente elongado.

*j.4) Determinación de conexiones múltiples.*

Los pasos precedentes del algoritmo están centrados en el establecimiento de relaciones de conexión entre pares de puntos extremos. En el paso presente se identifican las situaciones donde deben unirse tres o más puntos extremos mediante análisis del conjunto global de relaciones de conexión entre pares de puntos.

En el método desarrollado, se considera que tres o más puntos extremos deben ser conectados de forma simultánea, si y sólo si existen relaciones de conectividad entre cada par de elementos de este conjunto. Es decir, y en lenguaje de teoría de grafos, estos grupos corresponderán a los cliques (sub-conjuntos conexos completos máximos) del grafo inducido por el conjunto de puntos extremos con sus relaciones de conectividad asociadas. Los grupos de puntos extremos correspondientes a conexiones múltiples se determinan según el siguiente procedimiento:

- Determinación de los componentes conexos del grafo de conectividad mediante aplicación de un algoritmo de búsqueda en extensión, *breadth-first search* [Chartrand y Oellermann, 1993].

- Extracción jerárquica de los cliques de cada componente conexo en función de su tamaño.

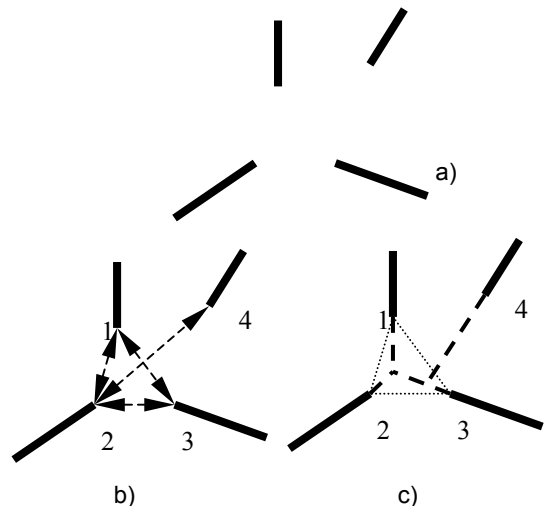

b) c)
**Figura 11**: Determinación de grupos de extremos de conexión múltiple. a) red inicial; b) relaciones de conectividad entre segmentos; c) grupo de conexión múltiple obtenido (extremos 1-2-3). El extremo 4, aislado, se une a la red en una fase posterior (ver texto).

En la figura 11 se muestra un conjunto de cuatro puntos extremos formando un componente conexo. En este conjunto, se encuentran los siguientes cliques:

- clique formado por los extremos 1-2-3.
- clique formado por los extremos 2-4.

De acuerdo al procedimiento jerárquico mencionado, se extrae el clique de mayor tamaño (puntos 1-2-3) para su conexión múltiple y se procede a analizar el resto del componente, que en este caso corresponde a un único elemento aislado, por lo que finaliza el proceso. Este extremo se conectará a la malla en una fase posterior del proceso (paso *j.6*).

*j.5) Trazado de las conexiones establecidas*

En este paso del proceso se unen físicamente, mediante segmentos de recta, los puntos extremos conectados lógicamente en los pasos anteriores. La conexión entre dos puntos extremos se realiza según el siguiente procedimiento:

- La conexión se realiza mediante trazado directo del segmento que une ambos extremos, si los últimos segmentos de ambos arcos son aproximadamente colineales.

- La conexión se realiza mediante prolongación del último segmento de cada arco, si estos no son colineales.

Las conexiones múltiples (involucrando tres o mas puntos extremos) se realizan mediante el siguiente procedimiento:

- Determinación de las intersecciones obtenidas mediante prolongación de los segmentos finales de cada par de los arcos a unir.

- Cálculo del baricentro de los puntos de intersección.

- Unión, mediante segmentos de recta, de cada punto extremo al baricentro calculado.

En la figura 12 se presentan un ejemplo de cada uno de los tipos de unión enumerados.

*j.6) Conexión de puntos extremos aislados*

En la última fase del proceso de conexión, se evalúa el coste de la conexión a la malla de puntos extremos aislados, mediante prolongación del último segmento de la aproximación lineal al arco. Esta conexión se realiza en caso de que el coste sea menor que un umbral. Este paso permite la conexión del punto extremo aislado de la figura 11.

k) *Almacenamiento en formato vectorial*

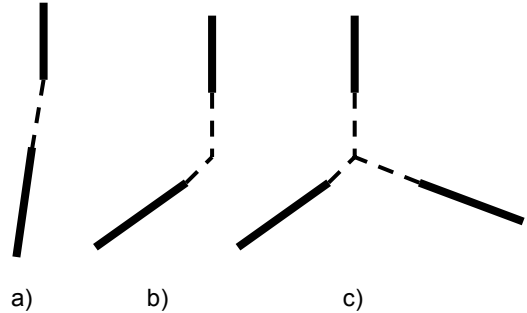

**Figura 12**: Trazado de las conexiones entre puntos extremos. a) Par de puntos extremos con arcos aproximadamente colineales; b) Par de puntos extremos con arcos no colineales; c) conexión múltiple.

Como último paso del procedimiento global, la red lineal obtenida se almacena en el formato vectorial de elección. De forma preliminar se ha utilizado para este propósito el formato DXF (Drawing eXchange Format) de Autodesk [Murray y van Ryper, 1996], dada su amplia difusión en el mercado.

**RESULTADOS OBTENIDOS**

De forma preliminar, el sistema descrito ha sido evaluado cualitativamente mediante inspección visual de los resultados obtenidos sobre una base de datos de mapas de distinta procedencia y escala, seleccionados específicamente para evaluar las prestaciones del método en zonas con alta densidad de información. Puede decirse que los resultados obtenidos sobre esta base de datos han sido globalmente de alta calidad, presentando en todos los casos un número reducido de zonas que requieren operaciones de retoque manual.

En la figura 13 se muestran un resultado de la extracción de la red de carreteras y la red fluvial sobre mapas de la base de datos utilizada.

**REFERENCIAS**

- ADAMS R., BISCHOF L. 1994. "Seeded Region Growing", *IEEE Transactions on Pattern Analysis and Machine Intelligence,* 16: 641-647.

- BORGEFORS, G. 1988. "Hierarchical Chamfer Matching: A parametric Edge Matching Algorithm", *IEEE Transactions on Pattern Analysis and Machine Intelligence*, 10: 849-865.

- CHARTRAND, G. OELLERMANN, O. "*Applied and Algorithmic Graph Theory*", Mc Graw-Hill, International Series in Pure and Applied Mathematics, 1993.

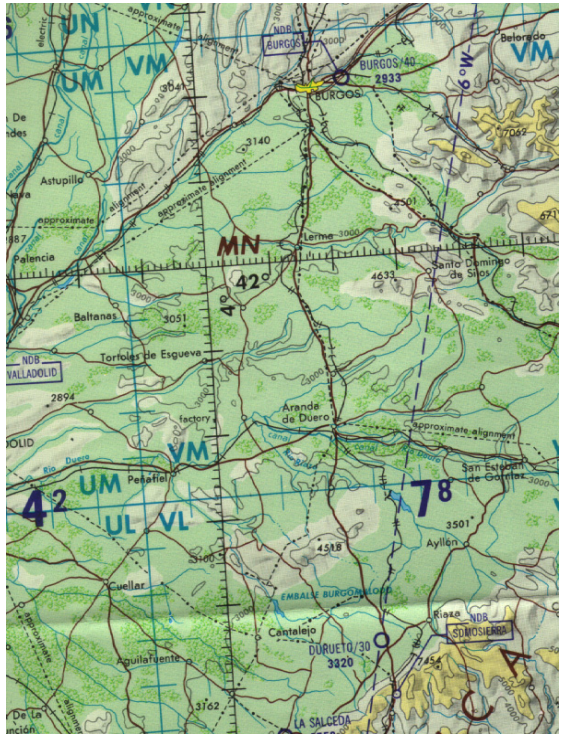

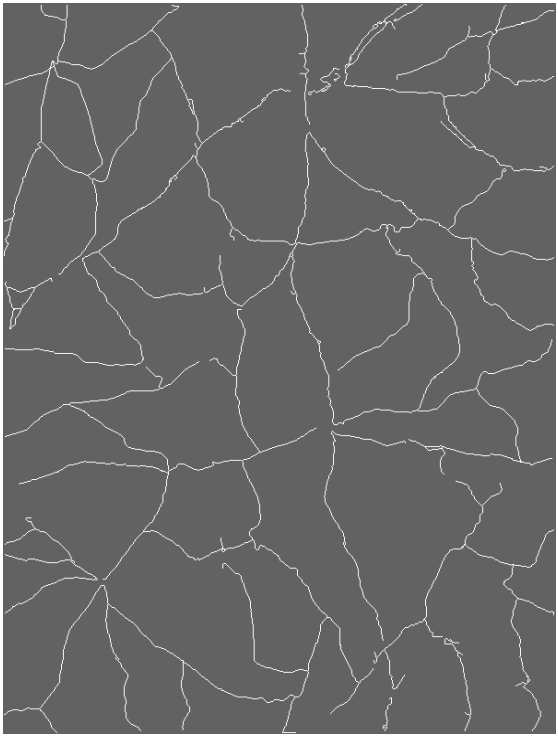

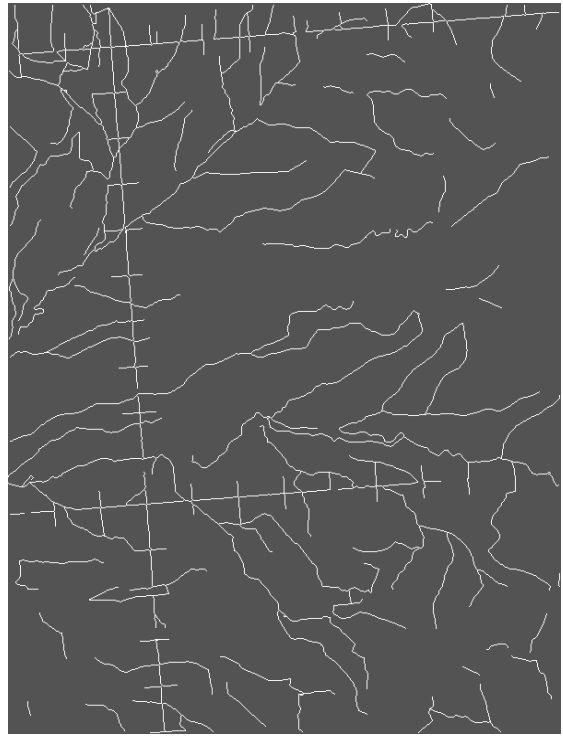

b)                                                        c)

**Figura 13**: a) mapa de entrada; b) red de carreteras; c) red fluvial.

- DUDA R., HART P. "*Pattern Classification and Scene Analysis*", Wiley -Interscience, 1973.

- FUERTES, J. LUCENA, M. PEREZ DE LA BLANCA, N. FDEZ-VALDIVIA, J. 1996. "A New Method of Segmenting Color Images", *VII National Symposium on Pattern Recognition and Image Analysis.*

- HARALICK, R. SHAPIRO, L. *"Computer and Robot Vision",* Vol. 1, Addison-Wesley, 1992.

- JAIN, A. "*Fundamentals of Digital Image Processing",* Prentice Hall, 1989.

- MURRAY, J. VAN RYPER, W. "*Encyclopedia of Graphics File Formats",* 2º Ed., O'Reilly and Associates, 1996.

- PRATT, W.K. "*Digital Image Processing*", 2º Ed., John Wiley & Sons, 1991.

- SANNITI DI BAJA, G. 1994, "Well-Shaped, Stable and Reversible Skeletons from the (3, 4)-Distance Tranform", *Journal of Visual Communication and Image Representation, 5:107-115.*